\documentclass[11pt,a4paper]{article}
\usepackage[hyperref]{acl2021}
\usepackage{times}
\usepackage{amssymb}
\usepackage{bm}
\usepackage{arydshln}
\usepackage{latexsym}
\usepackage{bbm}
\usepackage{blindtext}
\usepackage{graphicx}
\usepackage{amsmath,amssymb}
\usepackage{booktabs}
\usepackage{adjustbox}
\usepackage[toc,page]{appendix}

\usepackage{microtype}

\usepackage[normalem]{ulem}
\useunder{\uline}{\ul}{}

\usepackage{arydshln}  

\aclfinalcopy 

\newcommand{\method}[1]{\textsc{#1}\xspace}
\newcommand{\lastmention}{\method{LastMention}}
\newcommand{\glovemf}{\method{Glove$+$MF}}
\newcommand{\mf}{\method{MF}}
\newcommand{\bert}{\method{BERT}}
\newcommand{\bertmf}{\method{BERT$+$MF}}
\newcommand{\berttd}{\method{BERT$+$TD}}
\newcommand{\albert}{\method{ALBERT}}
\newcommand{\polyencoder}{\method{Poly-encoder}}
\newcommand{\polyinline}{\method{Poly-inline}}
\newcommand{\polybatch}{\method{Poly-batch}}
\newcommand{\multi}{\method{Multi}}
\newcommand{\multimf}{\method{Multi$+$MF}}

\title{Findings on Conversation Disentanglement}

\author{Rongxin Zhu, Jey Han Lau, Jianzhong Qi \\

        {School of Computing and Information System\rule[20pt]{0pt}{0pt}} \\
        {The University of Melbourne} \\
        rongxinz1@student.unimelb.edu.au, \{laujh, jianzhong.qi\}@unimelb.edu.au }

\date{}

\begin{document}
\maketitle
\begin{abstract}

Conversation disentanglement, the task to identify separate threads in conversations, is an important pre-processing step in multi-party conversational NLP applications such as conversational question answering and conversation summarization. Framing it as a utterance-to-utterance classification problem --- i.e.\ given an utterance of interest (UOI), find which past utterance it replies to --- we explore a number of transformer-based models and found that BERT in combination with handcrafted features remains a strong baseline. We then build a multi-task learning model that jointly learns utterance-to-utterance and utterance-to-thread classification. Observing that the ground truth label (past utterance) is in the top candidates when our model makes an error, we experiment with using bipartite graphs as a post-processing step to learn how to best match \textit{a set of UOIs} to past utterances. Experiments on the Ubuntu IRC dataset show that this approach has the potential to outperform the conventional greedy approach of simply selecting the highest probability candidate for each UOI independently, indicating a promising future research direction.

\end{abstract}

\section{Introduction}

\begin{figure}[t]
\centering
\includegraphics[width=8cm]{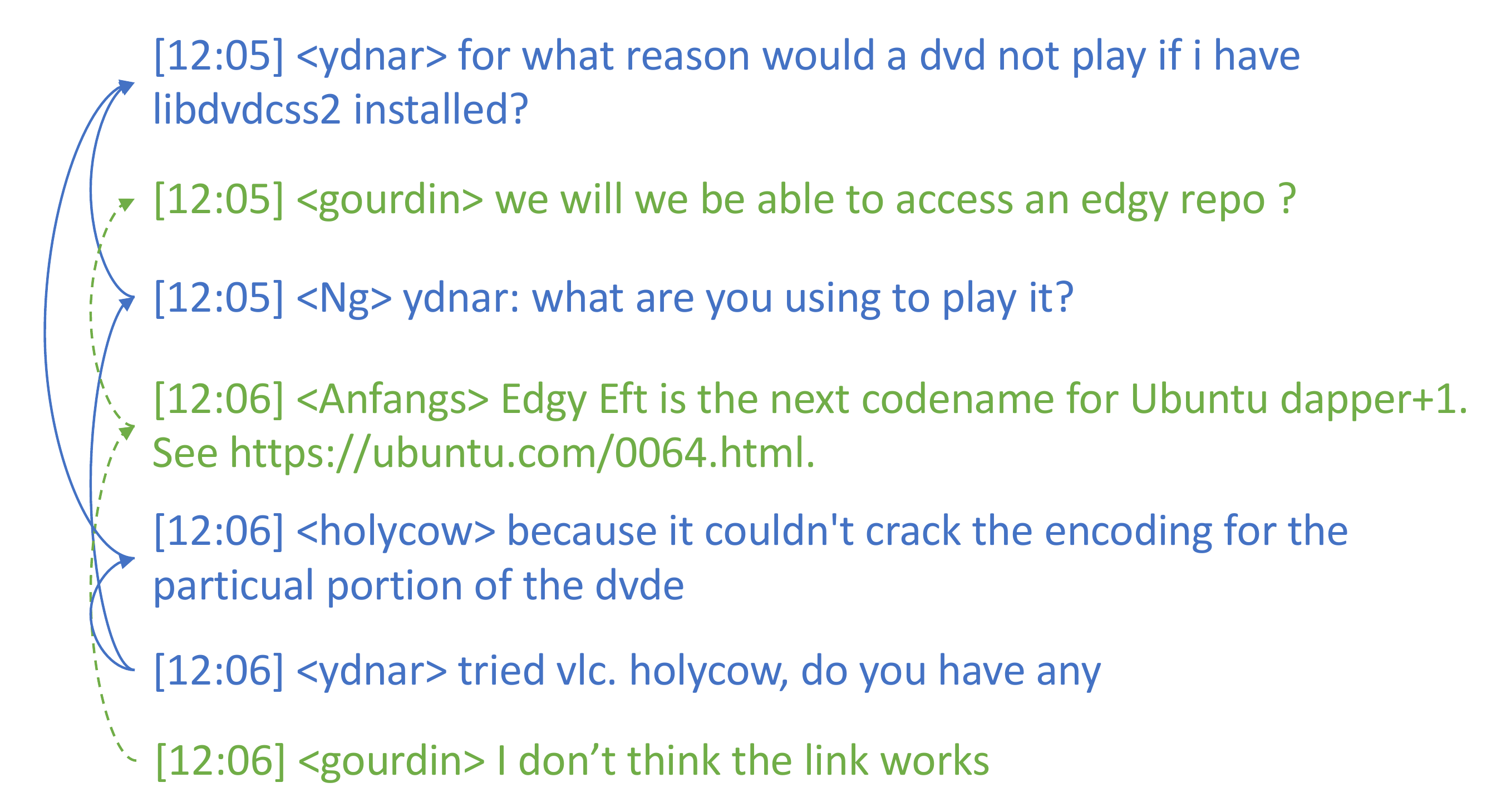}
\caption{Ubuntu IRC chat log sample sorted by time. Each arrow represents a directed reply-to relation. The two conversation threads are shown in blue and green.}
\label{fig:example}
\end{figure}

In public forums and chatrooms such as Reddit and Internet Relay Chat (IRC), there are often multiple conversations happening at the same time. Figure~\ref{fig:example} shows two threads of conversation (blue and green) running in parallel. \emph{Conversation disentanglement}, a task to identify separate threads among intertwined messages, is an 
essential preprocessing step for analysing entangled conversations in multi-party conversational applications such as question answering \cite{li_etal_2020_molweni} and response selection \cite{jia2020multi}. It is also useful in constructing datasets for dialogue system studies~\cite{lowe2015ubuntu}.

Previous studies address the conversation disentanglement task with two 
steps: \textit{link prediction} and \textit{clustering}. In link 
prediction, a confidence score is computed to predict a \textit{reply-to}
relation from an \emph{utterance of interest} (UOI) to a past utterance 
\cite{elsner-charniak-2008-talking0,zhu2020did}. In {clustering}, conversation threads are recovered based on the 
predicted confidence scores between utterance pairs.  The most popular 
clustering method uses a greedy approach to group UOIs linked with their 
best past utterances to create the threads 
\cite{kummerfeldetal2019large, zhu2020did}.

In \textit{link prediction}, the model that estimates the relevance 
between a pair of utterances plays an important role. To this end, we explore three transformer-based pretrained models: 
\bert~\cite{devlin_etal_2019_bert}, \albert~\cite{lan2019albert} and 
\polyencoder~\cite{humeau2019poly}. These variants are selected by 
considering performance, memory consumption and speed. We found that 
\bert in combination with handcrafted features remains a strong baseline.
Observing that utterances may be too short to contain sufficient 
information for disentanglement, we also build a multi-task learning 
model that learns to jointly link a UOI to a past utterance and a 
cluster of past utterances (i.e.\ the conversation threads).

For \textit{clustering}, we experiment with  bipartite graph matching 
algorithms that consider how to best link a set of UOIs to their top 
candidates, thereby producing globally more optimal clusters.  When the 
graph structure is known, we show that this approach 
substantially outperforms conventional greedy clustering method, 
although challenges remain on how to infer the graph structure.

To summarise:
\begin{itemize}
    \item We study different transformer-based models for conversation 
    disentanglement.
    \item We explore a multi-task conversation disentanglement framework 
    that jointly learns utterance-to-utterance and utterance-to-thread 
classification.
    \item We experiment with bipartite graphs for clustering utterances 
and found a promising future direction.
\end{itemize}

\section{Related Work}
Conversation disentanglement methods can be classified into two 
categories: (1) two-step methods and (2) end-to-end methods. 

In two-step methods, the first step is to measure the relations between 
utterance pairs, e.g., \textit{reply-to} relations \cite{zhu2020did, 
kummerfeldetal2019large} or \textit{same thread} relations 
\cite{elsner-charniak-2008-talking0, 
elsner-charniak-2010-disentangling}. Either feature-based models 
\cite{elsner-charniak-2008-talking0, elsner-charniak-2010-disentangling} 
or deep learning models \cite{kummerfeldetal2019large, zhu2020did} are 
used. Afterwards a clustering algorithm is applied to recover separate 
threads using results from the first step.  
\newcite{elsner-charniak-2008-talking0, 
elsner-charniak-2010-disentangling, elsner-charniak-2011-disentangling} 
use a greedy graph partition algorithm to assign an utterance $u$ to the 
thread of $u'$ which has the maximum relevance to $u$ among candidates 
if the score is larger than a threshold.  
\newcite{kummerfeldetal2019large, zhu2020did} use a greedy algorithm to 
recover threads following all reply-to relations independently 
identified for each utterance. \newcite{jiang2018learning} propose a 
graph connected component-based algorithm.

End-to-end methods construct threads incrementally by scanning through a 
chat log and either append the current utterance to an existing thread 
or create a new thread. \newcite{tan2019context} use a hierarchical LSTM 
model to obtain utterance representation and thread representation. 
\newcite{liu2020end} build a transition-based model that uses three LSTMs 
for utterance encoding, context encoding and thread state updating.

\section{Notations and Task Definition}
Given a chat log $U$ with $N$ utterances $\{u_1, u_2, \cdots, u_N\}$ in chronological order, the goal of conversation disentanglement is to obtain a set of disjoint threads $T = \{\mathcal{T}^1, \mathcal{T}^2, \cdots, \mathcal{T}^m\}$. Each thread $\mathcal{T}^l$ contains a collection of topically-coherent utterances. Utterance $u_i$ contains a list of $n_i$ tokens $w^i_1, w^i_2, \cdots, w^i_{n_i}$. 

The task can be framed as a \textit{reply-to relation identification} problem, where we aim to find the \emph{parent utterance} for every $u_i \in U$ \cite{kummerfeldetal2019large, zhu2020did}, i.e.,\ if an utterance $u_i$ replies to a (past) utterance $u_j$, $u_j$ is called the parent utterance of $u_i$. When all reply-to utterance pairs are identified, $T$ can be recovered unambiguously by following the reply-to relations.

Henceforth we call the target utterance $u_i$ an \textit{utterance of interest} (UOI). We use $u_i \rightarrow u_j$ to represent the reply-to relation from $u_i$ to $u_j$, where $u_j$ is the parent utterance of $u_i$. The reply-to relation is asymmetric, i.e., $u_i \rightarrow u_j$ and $u_j \rightarrow u_i$ do not hold at the same time. We use a \emph{candidate pool} $C_i$ to denote the set of candidate utterances from which the parent utterance is selected from.
Table~\ref{symbols_table} presents a summary of symbols/notations.

\begin{table}[t]
\small
\centering
\begin{tabular}{rl}
\toprule
Symbol               & Meaning                                                                \\
\midrule
$U$                    & A chat log with $N$ utterances                                                               \\
$T$                    & A set of disjoint threads in $U$                                                             \\
$\mathcal{T}$ & A thread in $T$                                                                              \\
$u_i$           & An utterance of interest                                                           \\
$u$                    & An utterance in a chat log                                                                 \\
$C_i$               & \begin{tabular}[c]{@{}l@{}}A candidate (parent) utterance pool for $u_i$\end{tabular} \\
$t_i$                 & The token sequence of $u_i$ with $n_i$ tokens                                                    \\
\bottomrule
\end{tabular}
\caption{\label{symbols_table} A summary of symbols/notations.}
\end{table}

\section{Dataset}
We conduct experiments on the Ubuntu IRC dataset~\cite{kummerfeldetal2019large}, which contains questions and answers about the Ubuntu system, as well as chit-chats from multiple participants. Table~\ref{dataset} shows the statistics in train, validation and test sets. The four columns are the number of chat logs, the number of annotated utterances, the number of threads and the average number of parents for each utterance.

\begin{table}[t]
\centering
\begin{adjustbox}{max width=\linewidth}
\begin{tabular}{rccccc}
\toprule
Split & Chat Logs & Ann. Utt & Threads & Avg. parent \\
\midrule
Train & 153       & 67463          & 17327     & 1.03              \\
Valid   & 10        & 2500           & 495       & 1.04              \\
Test  & 10        & 5000           & 964       & 1.04              \\
\bottomrule
\end{tabular}
\end{adjustbox}
\caption{\label{dataset} Statistics of training, validation and testing 
split of the Ubuntu IRC dataset. ``Ann. Utt'' is the number of annotated 
utterances. ``Avg. parent'' is the average number of parents of an 
utterance.} \end{table}

\section{Experiments}

We start with studying pairwise models that take as input a pair of utterances and decide whether a reply-to relation exists (Section~\ref{pairwise}). Then, we add dialogue history information into  consideration and study a {multi-task learning} model (Section~\ref{multi-task-learning}) built upon the  pairwise models. In Section~\ref{BGMCD}, we further investigate a globally-optimal approach based on {bipartite graph matching}, considering the top parent candidates of multiple UOIs together to help resolve conflicts in the utterance matches. 

\subsection{Pairwise Models}\label{pairwise}
To establish a baseline, we first study the effectiveness of
pairwise models that measure the confidence of a reply-to relation between an UOI and each candidate utterance independently without considering any past context (e.g., dialogue history). To find the parent utterance for $u_i$, we compute the relevance score $r_{ij}$ between $u_i$ and each $u_j \in C_i$:
\begin{equation}
r_{ij} = f(u_i, u_j, \bm{v_{ij}}), \forall\ u_j \in C_i
\end{equation}
where $f(\cdot)$ is the pairwise model and $\bm{v_{ij}}$ represents 
additional information describing the relationship between $u_i$ and 
$u_j$, such as manually defined features like time, user (name) mentions and  word overlaps. We use transformer-based models to automatically capture 
more complex semantic relationships between utterances pairs, such as 
question-answer relation and coreference resolution which cannot be 
modeled by features very well.

Following Kummerfeld et al.~(\citeyear{kummerfeldetal2019large}), we assume the parent utterance of a UOI to be within $k_c$ history utterances in the chat log, and we solve a $k_c$-way multi-class classification problem where $C_i$ contains exactly $k_c$ utterances $[u_{i-k_c+1}, \cdots, u_{i-1}, u_i]$. UOI $u_i$ is included in $C_i$ for detecting self-links, i.e.,\ an utterance that starts a new thread. The training loss is:
\begin{equation}
L_r = -\sum\limits_{i=1}^{N} \sum\limits_{j=1}^{k_c} {\mathbbm{1}[y_i=j]\log \ {p_{ij}} }
\end{equation}
where $\mathbbm{1}[y_i = j] = 1$ if $u_i \rightarrow u_j$ holds, and 0 otherwise; $p_{ij}$ is the normalized probability after applying softmax over $\{r_{ij}\}_{u_j \in C_i}$.

\subsubsection{Models}

We study the empirical performance of the following pairwise models. See more details of the models in Appendix~\ref{sec:appendix}.

\noindent \textbf{\lastmention}: A baseline model that links a UOI $u_i$ 
to the last utterance of the user directly mentioned by $u_i$. If $u_i$ 
does not contain a user mention, we link it to the immediately preceding 
utterance, i.e.,\ $u_{i-1}$.

\noindent \textbf{\glovemf}: Following \newcite{kummerfeldetal2019large}, this is a {feedforward neural network} (FFN) that uses the {max} and {mean} Glove~\cite{pennington2014glove} embeddings of a pair of utterances and some handcrafted features\footnote{See a full feature list in \newcite{kummerfeldetal2019large}.} including time difference between two utterances, direct user mention, word overlaps, etc.

\noindent \textbf{\mf}: An FFN model that uses only the handcrafted features in \glovemf. This model is designed to test the effectiveness of the handcrafted features.\footnote{Note that \mf is different from the manual features model in \newcite{kummerfeldetal2019large} which uses a linear model.}

\noindent \textbf{\bert~\cite{devlin_etal_2019_bert}}: A pretrained model based on transformer~\cite{vaswani2017attention} fine-tuned on our task. We follow the standard setup for sentence pair scoring in \bert by concatenating UOI $u_i$ and a candidate $u_j$ delimited by [SEP]. 

\noindent \textbf{\bertmf}: A BERT-based model that also incorporates the handcrafted features in \glovemf.

\noindent \textbf{\berttd}: A BERT-based model that uses the {t}ime {d}ifference between two utterances as the only manual feature, as preliminary experiments found that this is the most important feature.

\noindent \textbf{\albert~\cite{lan2019albert}}: A parameter-efficient 
\bert variant fine-tuned on our task.

\noindent \textbf{\polyencoder~\cite{humeau2019poly}}: A 
transformer-based model designed for fast training and inference by 
encoding query (UOI) and candidate separately.\footnote{It is worthwhile 
to note that \polyencoder showed strong performance on a related 
task, {next utterance selection}, which aims to choose the correct 
future utterance, but with two key differences: (1) their UOI 
incorporates the dialogue history which provides more context; (2) they 
randomly sample negative examples to create the candidates, while we use 
$k_c$ past utterances as candidates, which makes the next utterance 
selection task arguably an easier task.}
We use \polyencoder in two settings: \polybatch where the labels of 
UOIs in a batch is used as the shared candidate pool to reduce 
computation overhead, and \polyinline where each query has its own 
candidate pool similar to the other models.

\begin{table*}[t]
\centering
\small
\begin{tabular}{lccccccccc}
\toprule
                   & \multicolumn{3}{c}{Link Prediction} & \multicolumn{3}{c}{Ranking} & \multicolumn{3}{c}{Clustering} \\
Model              & Precision     & Recall    & F1      & R@1     & R@5     & R@10    & 1-1      & VI       & F        \\ \midrule
Last Mention       & 37.1          & 35.7      & 36.4    & -       & -       & -       & 21.4     & 60.5     & 4.0      \\ 
\glovemf & 71.5          & 68.9      & 70.1    & 70.2    & 95.8    & 98.6    & 76.1     & 91.5     & 34.0     \\
\mf         & 71.1          & 68.5      & 69.8    & 70.2    & 94.0    & 97.3    & 75.0     & 91.3     & 31.5     \\ \hdashline
\polybatch         & 39.3          & 37.9      & 38.6    & 40.8    & 69.8    & 80.8    & 52.3     & 80.8     & 9.8      \\
\polyinline        & 42.2          & 40.7      & 41.4    & 42.8    & 70.8    & 81.3    & 62.0     & 84.4     & 13.6     \\
\albert             & 46.1          & 44.4      & 45.3    & 46.8    & 77.3    & 88.4    & 68.6     & 87.9     & 22.4     \\ 
\bert               & 48.2          & 46.4      & 47.3    & 48.8    & 75.4    & 84.7    & 74.3     & 89.3     & 26.3     \\ \hdashline
\berttd          & 67.9          & 65.4      & 66.6    & 66.9    & 90.6    & 95.3    & 76.0     & 91.1     & 34.9     \\
\bertmf & \textbf{73.9} & \textbf{71.3} & \textbf{72.6} & \textbf{73.9} & \textbf{95.8} & \textbf{98.6} & \textbf{77.0} & \textbf{92.0} & \textbf{40.9} \\ \bottomrule
\end{tabular}
\caption{\label{context-free models} Results of pairwise models. \textit{Ranking metrics} are not applicable to \textit{Last Mention}. Best scores are \textbf{bold}.}
\end{table*}

\subsubsection{Results}\label{context_free_experiments}
\paragraph{Evaluation Metrics}
We measure the model performance in three aspects: (1) the \textit{link 
prediction} metrics measure the precision, recall and F1 scores of the 
predicted reply-to relations; (2) the \textit{clustering} metrics 
include variation information (VI, \cite{meilua2007comparing}), 
one-to-one Overlap (1-1, \cite{elsner-charniak-2008-talking0}) and exact 
match F1; these evaluate the quality of the recovered 
threads;\footnote{Exact Match F1 is calculated based on the number of 
recovered threads that perfectly match the ground truth ones (ignoring the 
ground truth threads with only one utterance).} and (3) the 
\textit{ranking} metrics Recall@$k$ ($k = \{1, 5, 10\}$) assess whether 
the ground truth parent utterance $u_j$ is among the top-$k$ 
candidates.\footnote{E.g.,\ if $u_j$ is in the top-$5$ candidates, 
recall@$5$ = 1.}

\paragraph{Dataset construction}
In training and validation, we set $C_i$ to contain exactly one parent 
utterance of an UOI $u_i$. We observe that $98.5\%$ of the UOIs in the 
training data reply to a parent utterance within the 50 latest 
utterances and so we set $k_c=50$ (i.e., $|C_i| = 50$).
We discard training samples that do no contain the parent utterance of 
an UOI under this setting ($1.5\%$ in the training data).
If there are more than one parent utterances in $C_i$ ($2.5\%$ in training data), we take the latest parent utterance of $u_i$ as the target ``label''. 

We do not impose these requirements in testing and so do not manipulate 
the test data.

\paragraph{Model configuration}
We clip both UOI $u_i$ and a candidate $u_j$ to at most 60 tokens.  
$|\bm{v_{ij}}|$ (manual feature dimension) $ = 77$ in \bertmf. In 
\berttd, $|\bm{v_{ij}}| = 6$.  The dimensionality of word embeddings in 
\mf is 
50. All BERT-based models use the ``bert-base-uncased'' pretrained model.  
    The batch size for \polyinline, \bert, \berttd and \bertmf is 
64.\footnote{Actual batch size is 4 with a gradient accumulation of  
   16.} The batch sizes of \polybatch and \albert are 
96 and 256 respectively. We tune the batch size, the number of layers, and the 
   hidden size in \bertmf and \berttd according to recall@1 on the 
validation set.

\paragraph{Results and discussions}
 Table~\ref{context-free models} shows that \lastmention is worse 
than all other models, indicating that direct user mentions are not 
sufficient for disentanglement. The manual features model (\mf) has very 
strong results, outperforming transformer-based  models 
(\bert, \albert and \polyencoder) by a large margin, suggesting that the 
manual features are very effective.

The overall best model across all metrics is \bertmf. Comparing \bertmf to \bert, we see a large improvement when we incorporate the manual features. Interestingly though, most of the improvement appears to come from the time difference feature (\bertmf vs.\ \berttd).

Looking at \bert and \polyinline, we see that the attention between 
words in BERT is helpful to capture the semantics between utterance 
pairs better, because the only difference between them is that 
\polyinline encodes two utterances separately first and uses  additional 
attention layers to compute the final relevance score.

The performance gap between \polybatch and \polyinline shows that the 
\textit{batch mode}~\cite{humeau2019poly} strategy has a negative impact 
on the prediction accuracy. This is attributed to the difference in 
terms of training and testing behaviour, as at test time we  
predict links similar to the inline mode (using past $k_c$ utterances 
as candidates).

The GPU memory consumption and speed of transformer-based models are 
shown in Table~\ref{table:transformers}. \polybatch is the most memory 
efficient and fastest model, suggesting that it is a competitive model in 
real-world applications where speed and efficiency is paramount.

\begin{table}[t]
\centering
\begin{adjustbox}{max width=\linewidth}
\begin{tabular}{ccc}
\toprule
Model & GPU Mem (GB) & Speed (ins/s) \\
\midrule
\bert                      & 18.7                 & 9.4                  \\
\albert                    & 14.6                 & 9.4                  \\
\polyinline               & 9.9                  & 16.8                 \\
\polybatch                & 5.1                  & 36.4                 \\
\bottomrule
\end{tabular}
\end{adjustbox}
\caption{\label{table:transformers} GPU memory consumption and speed of transformer-based models. GPU Mem (GB) shows the peak GPU memory consumption in GB during training. Speed (ins/s) is the number of instances processed per second during training. All experiments are conducted on a single NVIDIA V100 GPU (32G) with automatic mixed precision turned on and a batch size of 4.}
\end{table}

\subsection{Context Expansion by Thread Classification}\label{multi-task-learning}
The inherent limitation of the pairwise models is that they ignore the dialogue history of a candidate utterance. Intuitively, if the prior utterances from the same thread of candidate utterance $u_j$ is known, it will provide more context when computing the relevance scores.
 However, the threads of candidate utterances have to be inferred, which could be noisy. Furthermore,  the high GPU memory consumption of transformer-based 
models renders using a long dialogue history impractical. 

To address the issues above, we propose a multi-task learning framework  that (1) considers the dialogue history in a memory efficient manner and (2) does not introduce noise at test time. 

Specifically, we maintain a candidate thread pool with $k_t$ threads. A thread that contains multiple candidates would only be included once.

This alleviates some of the memory burden, not to mention that  $k_t$ is much smaller than $|C_i|$. For the second issue, we train a shared \bert model that does {reply-to relation identification} and {thread classification} jointly, and during training we use the ground truth threads but at test time we only perform reply-to relation identification, avoiding the use of potentially noisy (predicted) threads.

\begin{figure*}[ht]
\centering
\includegraphics[width=15cm]{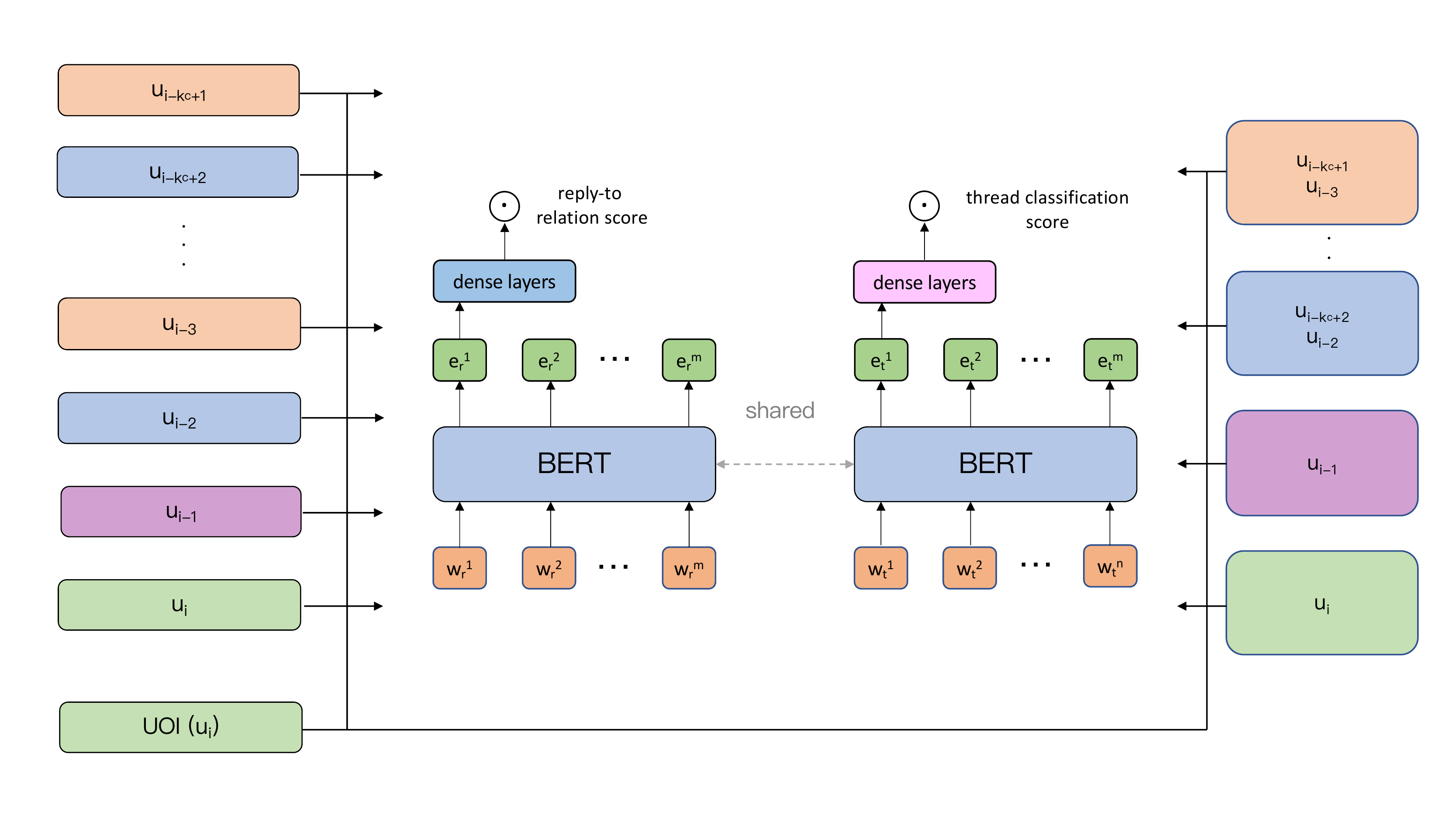}
\caption{The architecture of the multi-task learning framework. On the left side, we use a \bert model with additional dense layers to calculate the relevance score between a UOI and each candidate utterance for reply-to relation identification. On the right side, we use the same BERT model but  different dense layers on the top to calculate the relevance scores between the UOI and each candidate thread for thread classification.}
\label{multi-task-archi}
\end{figure*}

\subsubsection{Model Architecture}
The model consists of a shared \bert module and separate linear layers 
for {reply-to relation identification} and {thread classification}. As 
shown in Figure~\ref{multi-task-archi}, given $u_i$, we compute its 
relevance score $s^r_{ij}$ to every candidate utterances in utterance 
candidate pool $C_i$ and relevance score $s^t_{il}$ to every thread in 
thread candidate pool $T^c_i$. We aim to minimize the following loss 
function during model training:
\begin{equation}\label{eq:multi_task}
\begin{array}{l}
    L = - \Big( \sum\limits_{i=1}^{N} \sum\limits_{j=1}^{k_c} {\mathbbm{1}(y_r=j)\log s^r_{ij}} \\ 
       \quad \quad + \alpha \sum\limits_{i=1}^{N} \sum\limits_{l=1}^{k_t} {\mathbbm{1}(y_t=l)\log s^t_{il}}\Big)
\end{array}
\end{equation}
where $\mathbbm{1}(y_r=j)$ is 1 if $u_j$ is the parent utterance of 
$u_i$, and 0 otherwise. Similarly, $\mathbbm{1}(y_t=l)$ tests whether 
$u_i$ belongs to thread $T^c_i$. Hyper-parameter $\alpha$ is used to 
balance the importance of the two loss components.

\begin{table*}[t]
\centering
\small
\begin{adjustbox}{max width=\linewidth}
\begin{tabular}{@{}lccccccccc@{}}
\toprule
                                                  & \multicolumn{3}{c}{Link Prediction} & \multicolumn{3}{c}{Ranking} & \multicolumn{3}{c}{Clustering} \\ 
Model                                             & Precision     & Recall    & F1      & R@1     & R@5     & R@10    & 1-1      & VI       & F        \\ \midrule
\bert                                              & 48.2          & 46.4      & 47.3    & 48.8    & 75.4    & 84.7    & 74.3     & 89.3     & 26.3     \\
\bertmf                                        & 73.9          & 71.3      & 72.6    & 73.9    & 95.8    & 98.6    & 77.0     & 92.0     & 40.9     \\ \midrule
\multi ($\alpha=1$)       & 65.6          & 63.2      & 64.4    & 66.7    & 91.8    & 95.6    & 64.6     & 87.7     & 24.3     \\
\multi ($\alpha=5$)                                & 66.9          & 64.5      & 65.7    & 65.4    & 91.8    & 95.6    & 68.7     & 88.8     & 27.4     \\
\multi ($\alpha=10$)                               & 65.2          & 62.9      & 64.0    & 64.4    & 91.4    & 95.6    & 70.3     & 89.5     & 28.1     \\
\multi ($\alpha=20$)                               & 64.7          & 62.4      & 63.5    & 63.9    & 91.0    & 95.0    & 68.3     & 88.8     & 26.7     \\ \midrule
\multimf ($\alpha=1$) & 72.8          & 70.2      & 71.5    & 71.9    & 94.0    & 96.4    & 76.3     & 91.8     & 36.1     \\
\multimf ($\alpha=5$)                          & 73.3          & 70.7      & 72.0    & 72.4    & 94.0    & 96.5    & 72.8     & 90.8     & 33.1     \\
\multimf ($\alpha=10$)                         & 72.2          & 69.6      & 70.8    & 70.4    & 93.4    & 96.4    & 71.8     & 90.2     & 29.9     \\
\multimf ($\alpha=20$)                         & 70.8          & 68.2      & 69.5    & 69.4    & 93.4    & 97.3    & 73.2     & 90.6     & 28.6     \\ \bottomrule
\end{tabular}
\end{adjustbox}
\caption{\label{table:multi_task} Results of multi-task learning model.}
\end{table*}

\paragraph{Relevance score computation}

We compute the utterance relevance score $s^r_{ij}$ between UOI $u_i$ 
and each candidate utterance $u_j \in C_i$ in the same way as  the \bert model shown in Section~\ref{pairwise}.

For thread classification, we consider a pool containing $k_t$ threads 
before $u_i$, including a special thread $\{u_i\}$ for the case 
where  $u_i$ starts a new thread. The score $s^t_{il}$ between $u_i$ and 
thread $T_l$ is computed using the shared \bert, following the format used by 
\newcite{ghosal2020utterance}:
\begin{align*}
\big[& [\mbox{CLS}],w_1^1, \cdots w_{n_1}^1, w_1^2, \cdots w_{n_2}^2, w_1^k \cdots w_{n_k}^k,\\
& [\mbox{SEP}],w^i_1,\cdots w^i_{n_i} [\mbox{SEP}] \big]
\end{align*}
where $w^p_q$ is the $q$-th token of the $p$-th utterance in $T_l$, and $w^i_m$ is the $m$-th token of $u_i$. We take the embedding of $[\mbox{CLS}]$ and use another linear layer to compute the final score.

\subsubsection{Results and Discussion}
For reply-to relation identification, we use the same configuration 
described in Section \ref{context_free_experiments}. For thread 
classification, we consider $k_t = 10$ thread candidates. Each thread is 
represented by (at most) five latest utterances. The maximum number of 
tokens in $T_l$ and $t_i$ are $360$ and $60$, respectively.  We train 
the model using Adamax optimizer with learning rate $5 \times 10^{-5}$ 
and batch size 64. As before we use ``bert-base-uncased'' as the 
pretrained model.

As Table~\ref{table:multi_task} shows, incorporating an additional thread classification loss (``\multi ($\alpha = k$)'' models) improves link prediction substantially compared to \bert, showing that the thread classification objective provides complementary information to the reply-to relation identification task. Interestingly, when $\alpha$ increases from $5$ to $10$, both the link prediction and ranking metrics drop, suggesting that it is important not to over-emphasize thread classification, since it is not used at test time.

Adding thread classification when we have manual features (\multimf vs.\ \bertmf), however, does not seem to help, further reinforcing the effectiveness of these features in the dataset. That said,  
in situations/datasets where these manual features are not available, e.g.\ Movie Dialogue Dataset~\cite{liu2020end}, our multi-task learning framework could be useful.

\subsection{Bipartite Graph Matching for Conversation Disentanglement}\label{BGMCD}

\begin{figure}[t]
\centering
\includegraphics[width=8cm]{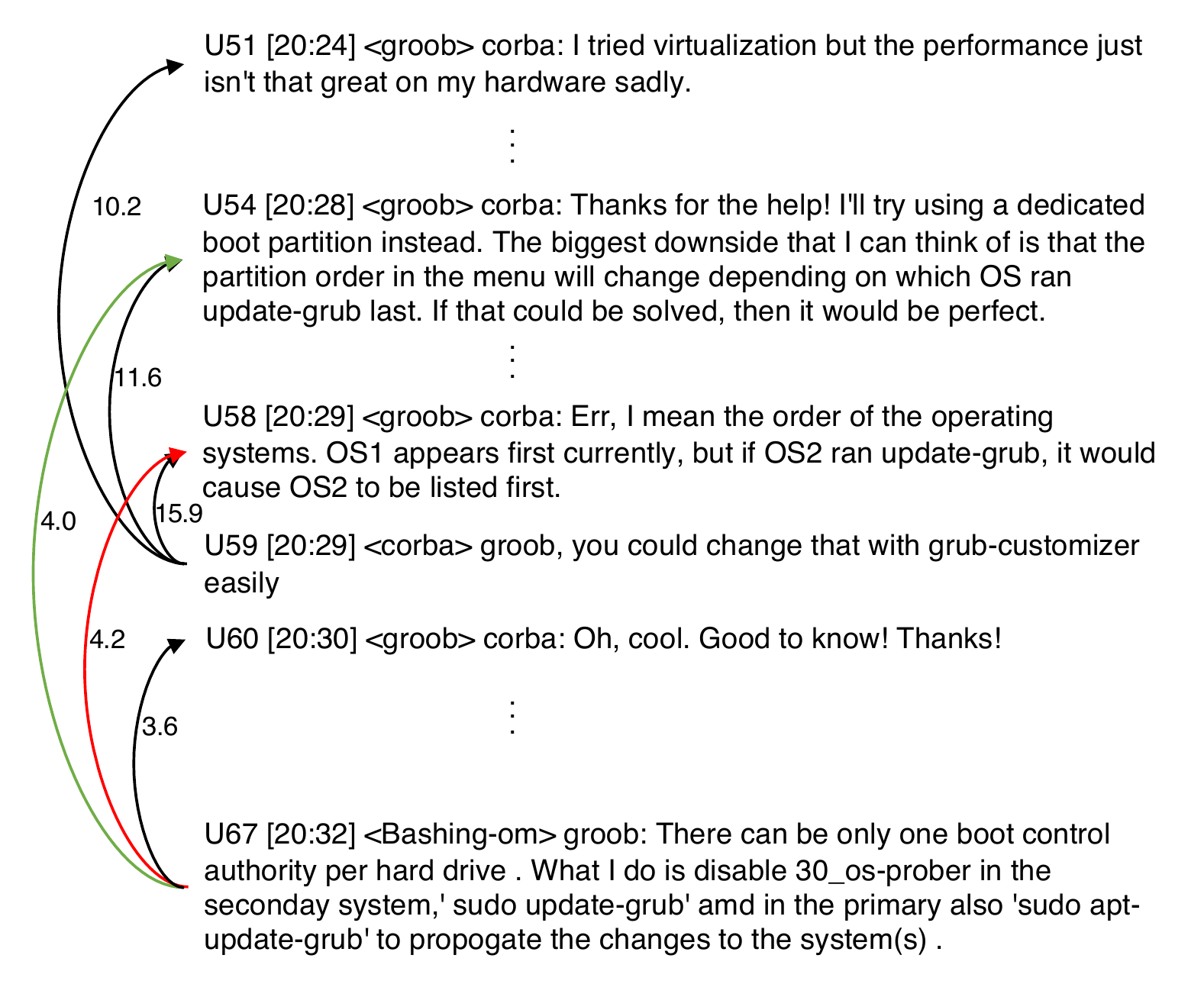}
\caption{An example showing the difference between the greedy approach and the global decoding. Consider identifying the parent utterances of $u_{59}$ and $u_{67}$. Each utterance contains ID (e.g., $u_{51}$), timestamp, user name and content. Both $u_{59}$ and $u_{67}$ have three candidates. The pairwise scores are labelled to the links, indicating the confidence of potential \textit{reply-to} relations. The red link denotes the identified \textit{reply-to} relation for $u_{67}$ using the greedy approach, and the green link is the result of a global decoding algorithm.}
\label{fig:greedy_vs_global}
\end{figure}

After we have obtained the pairwise utterance relevance scores for every UOI, we need to link the candidate utterances with the UOIs to recover the threads. A greedy approach would use all reply-to relations that have been identified \emph{independently} for each UOI to create the threads. As shown in Figure~\ref{fig:greedy_vs_global}, the \textit{reply-to} relations for $u_{67}$ and $u_{59}$ using greedy approach are $\{u_{67} \rightarrow u_{58}, u_{59} \rightarrow u_{58}\}$.

With such an approach, we observe that: (1) some candidates receive more responses than they should (based on ground truth labels); and (2) many UOIs choose the same candidate. Given the fact that over $95\%$ of the UOIs' parents are within the top-5 candidates in \bertmf (R@5 in Table~\ref{context-free models}), we explore whether it is possible to get better matches if we constrain the maximum number of reply links each candidate receives  and perform the linking of UOIs to their parent utterances together. In situations where  
a UOI $u_i$'s top-1 candidate utterance $u_j$ has a relevant score that is just marginally higher than other candidates but  $u_j$ is a strong candidate utterance for other UOIs, we may want to link $u_j$ with the other UOIs instead of $u_i$. Using Figure~\ref{fig:greedy_vs_global} as example, if $u_{58}$ can only receive one response, then $u_{67}$ should link to the second best candidate $u_{54}$ as its parent instead of $u_{58}$.

Based on this intuition, we explore using bipartite algorithms that treat the identification of all reply-to relations within a chat log as a \emph{maximum-weight matching}~\cite{gerards1995matching} problem on a bipartite graph. Note that this step is a post-processing step that can be applied to technically any pairwise utterance scoring models.

\subsubsection{Graph Construction}
Given a chat log $U$, we build a bipartite graph $G = \langle V, E, W \rangle$ where $V$ is the set of nodes, $E$ is the set of edges, and $W$ is the set of edge weights. Set $V$ consists of two subsets $V_l$ and $V_r$ representing two disjoint subsets of nodes of a bipartite. Subset $V_l=\{v^l_i\}_{i=1}^N$ represents the set of UOIs, i.e., each node $v^l_i$ corresponds to a UOI  $u_i$. Subset $V_r$ represents the set of candidate utterances. Note that some UOIs may be candidate utterances of other UOIs. Such an utterance will have both a node in $V_l$ and a node in $V_r$.

Some utterances may receive more than one reply, i.e.,\ multiple nodes in $V_l$ may link to the same node in $V_r$. This violates the standard assumption of a bipartite matching problem, where every node in $V_r$ will only be matched with at most one node in $V_l$. To address this issue, we duplicate nodes in $V_r$. 
Let $\delta(u_j)$ denotes the number of replies $u_j$ receives, 
then $u_j$ is represented by $\delta(u_j)$ nodes in $V_r$. Now $V_r = \bigcup_{j=1}^N{S(u_j)}$, where $S(u_j)$ is a set of duplicated nodes $\{v^r_{j,1}, v^r_{j,2}, \cdots v^r_{j, \delta(u_j)}\}$ for $u_j$.

\begin{figure}[t]
\centering
\includegraphics[width=7.9cm]{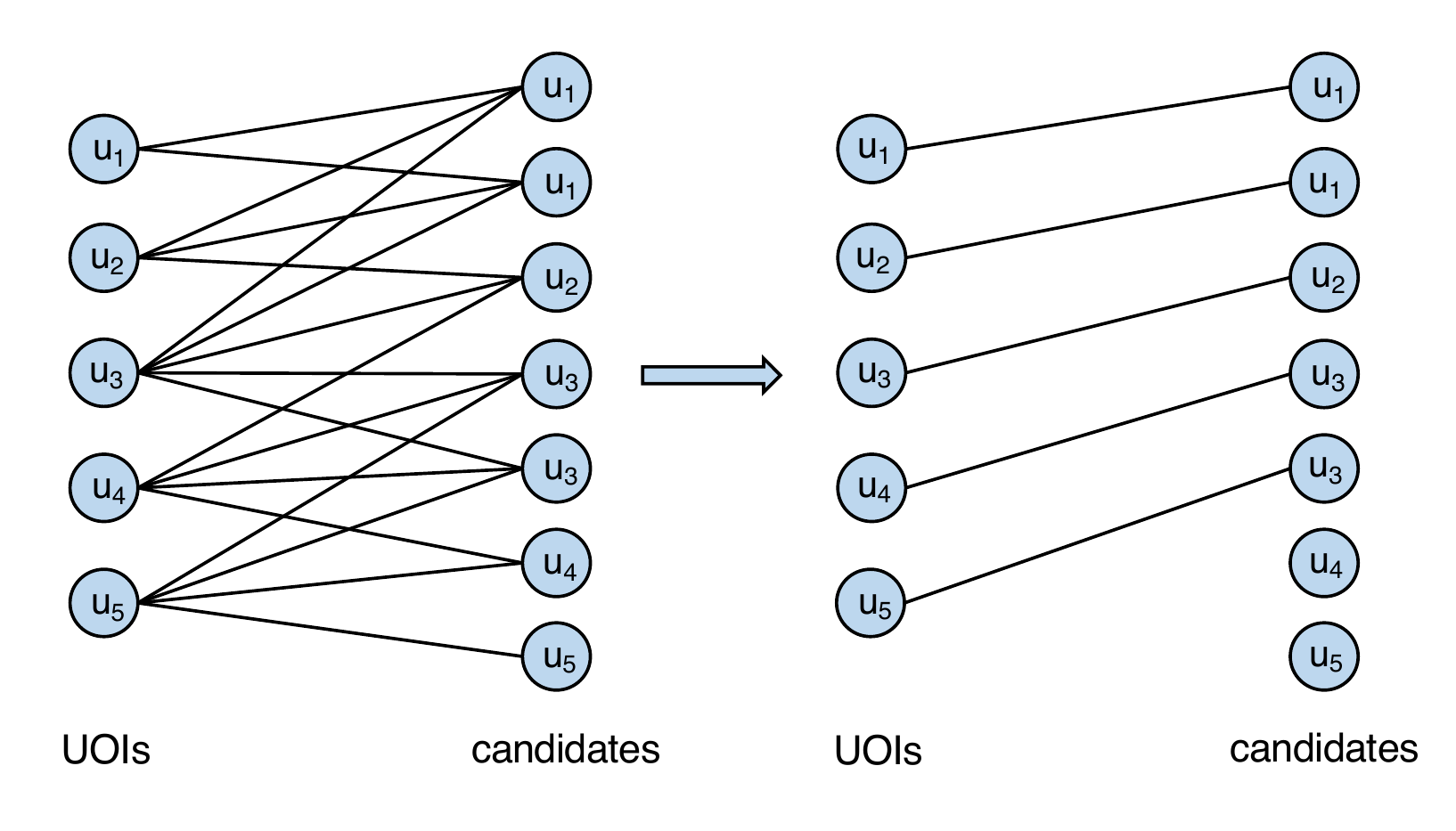}
\caption{The left figure is an example bipartite graph built from a chat log with 5 UOIs. Each UOI $u_i$ has $k_c=3$ candidates $\{u_{i-2}, u_{i-1}, u_i\}$, except the first $k_c - 1$ UOIs ($u_1$ and $u_2$). Utterances $u_1$ and $u_3$ are duplicated twice because they receive 2 replies. The corresponding disentangled chat log is shown on the right figure with the following reply-to relations: $\{u_1 \rightarrow u_1, u_2 \rightarrow u_1, u_3 \rightarrow u_2, u_4 \rightarrow u_3, u_5 \rightarrow u_3 \}$.}
\label{fig:bipartite_graph}
\end{figure}

Sets $E$ and $W$ are constructed based on the pairwise relevance scores obtained from the link prediction phase. Specifically, $E = \bigcup_{i=1}^{N}{R(u_i)}$ where $R(u_i)$ is the set of edges between $u_i$ and all its $k_c$ candidates: $\bigcup_{m=1}^{k_c}\{ \langle v^l_i, v_m \rangle \}_{v_m \in S(u_m)}$. For each UOI-candidate pair $( u_i, u_j )$, if $\delta(u_j) > 0$, a set of edges $\{\langle v^l_i, v^r_{j,k} \rangle \}_{k=1}^{\delta(u_j)}$ are constructed, each with weight $w(i,j)$, which is the relevance score between $u_i$ and $u_j$. An example bipartite graph is shown on the left side of Figure~\ref{fig:bipartite_graph}.

\subsubsection{Integer Programming Formulation}
Given the bipartite formulation above, we solve the 
 conversation disentanglement problem as a maximum-weight bipartite matching problem, which is formulated as the following constrained optimization problem:

\begin{equation}\label{eq:bgmcd}
\begin{array}{ll@{}ll}
                & \text{max} \displaystyle\sum\limits_{\langle v_i,v_j \rangle \in E} x(i,j) \cdot w(i,j)  & \\
    \text{s.t.} &  & &\\ 
            & \displaystyle\sum\limits_{v_l \in neighbors(v_i)}   x(i,l)  = 1, & \forall v_i \in V_l \\
            & \displaystyle\sum\limits_{v_p \in neighbors(v_j)}   x(p,j) \leq 1, & \forall v_j \in V_r \\
            & x(i,j) \in \{0, 1\}
\end{array}
\end{equation}

Here, $neighbors(v_x)$ is the set of adjacent nodes of $v_x$ (i.e., nodes directly connected to $v_x$) in $G$. For each edge in $G$, we have a variable $x(i,j)$, which takes value 1 if we include the edge $\langle v_i,v_j \rangle$ in the final matched bipartite, and 0 otherwise. Intuitively, we are choosing a subsect of $E$ to maximize the total weight of the chosen edges, given the constraints that (1) each node in set $V_l$ is connected to exactly one edge (each UOI has exactly one parent); and 
(2) each node in $V_r$ is connected to at most one edge.

\subsubsection{Node Frequency Estimation in $V_r$}

Since the number of replies received by an utterance $u_j$, i.e., $\delta(u_j)$, is unknown at test time, we estimate $\delta(u_j)$ for each candidate utterance $u_j$. 
We experiment with two different estimation strategies: {heuristics method} and {regression model}.

In the heuristics method, we estimate $\delta(u_j)$ based on the total relevance scores accumulated by $u_j$ from all UOIs, using the following equation:
\begin{align*}
    {r_{ij}}' &= \frac{\exp(r_{ij})}{\sum_{u_k \in C_i}{\exp(r_{ik})}}\\
    S_j &= \sum\limits_i {r_{ij}}'\\
\hat{\delta}(u_j) &= \text{RND}(\alpha S_j + \beta)
\end{align*}
where  $\hat{\delta}(u_j)$ is the estimation, $\text{RND}$ is the $round(\cdot)$ function, and $\alpha$ and $\beta$ are scaling parameters. 

In the regression model, we train an FFN to predict $\delta(u_j)$ using {mean squared error} as the training loss. The features are normalized scores of $u_j$ from all UOIs, as well as the sum of those scores. We also include textual features using BERT (based on the [CLS] vector), denoted as BERT$+$FFN. We use the same $\text{RND}$ function to obtain an integer from the prediction of the regression models.

\subsubsection{Experiments and Discussion}
\begin{table}[t]
\centering
\begin{adjustbox}{max width=\linewidth}
\begin{tabular}{lccc}
\toprule
           & Precision & Recall & F1   \\ \midrule
{Oracle}  & 88.4      & 85.2   & 86.8 \\
Rule-Based & 73.7      & 70.9   & 72.3 \\
FFN        & 73.8      & 71.0   & 72.3 \\ 
BERT$+$FFN   & 72.9      & 70.3   & 71.5 \\ \bottomrule
\end{tabular}
\end{adjustbox}
\caption{\label{cheat-bgmcd} Link prediction results using bipartite matching. \textit{Oracle} is a model that uses ground truth node frequencies for $V_r$.}
\end{table}

We obtain the performance upper bound by solving the maximum weight bipartite matching problem using the ground truth node frequencies for all nodes in $V_r$. This approach is denoted as ``Oracle'' in Table~\ref{cheat-bgmcd}. 
We found that when node frequencies are known, bipartite matching significantly outperforms the best greedy methods (F1 score 86.8 vs.\ 72.6 of \bertmf in Table \ref{context-free models}).  

When using estimated node frequencies, the heuristics method and FFN achieve very similar results, and BERT$+$FFN is worse than both. Unfortunately, these results are all far from Oracle, and they are ultimately marginally worse than \bertmf (72.6; Table \ref{context-free models}). Overall, our results suggest that there is much potential of using bipartite matching for creating the threads, but that there is still work to be done to design a more effective method for estimating the node frequencies.

\section{Conclusion}
In this paper, we frame conversation disentanglement as a task to identify the past utterance(s) that each utterance of interest (UOI) replies to, and conduct various experiments to explore the task. We first experiment with transformer-based models, and found that BERT combined with manual features is still a strong baseline. Next we propose a multi-task learning model to incorporate dialogue history into BERT, and show that the method is effective especially when manual features are not available. Based on the observation that most utterances' parents are in the top-ranked candidates when there are errors, we experiment with bipartite graph matching that matches a set of UOIs and candidates together to produce globally more optimal clusters. The algorithm has the potential to outperform standard greedy approach, indicating a promising future research direction.

\section{Acknowledgement}
We would like to thank the anonymous reviewers for their helpful comments.

\bibliographystyle{acl_natbib}
\bibliography{anthology,acl2021}

\section{Appendix}
\label{sec:appendix}

\subsection{Models}
\paragraph{\bert}: The pairwise score is computed as follows:
\begin{equation}\label{eq:bert}
\begin{array}{l}
    [\bm{e_1}, \bm{e_2}, \cdots, \bm{e_{m}}] = \bert(concat(t_i, t_j)) \\
    \bm{e} = agg([\bm{e_1}, \bm{e_2}, \cdots, \bm{e_{m}}]) \\
    r_{ij} = \bm{W} \bm{e} + \bm{b}
\end{array}
\end{equation}

Here, $concat(t_i, t_j)$ means to concatenate the two sub-word sequences $t_i$ and $t_j$ corresponding to $u_i$ and $u_j$ into a single sequence  $\big[[\mbox{CLS}],w^i_1,\cdots,w^i_{n_i},[\mbox{SEP}],w^j_1,\cdots,w^j_{n_j},[\mbox{SEP}]\big]$, where $[\mbox{CLS}]$ is a special beginning token and $[\mbox{SEP}]$ is a separation token. Denote the number of tokens in this sequence by $m$. 
Then, $\bm{e_k} \in \mathbbm{R}^{d_{BERT}}$ is the encoded embedding of the $k$-{th} ($k \le m$) token in $t_{ij}$. Following \cite{devlin_etal_2019_bert}, we use the encoded embedding of $\mbox{[CLS]}$ as the aggregated representation of $u_i$ and $u_j$. Another linear layer is applied to obtain score $r_{ij} \in \mathbbm{R}$ using learnable  parameters $\bm{W} \in \mathbbm{R}^{1 \times d_{BERT}}$ and $\bm{b} \in \mathbbm{R}$.

\subsection{\bertmf}
We obtain the encoded embedding of [CLS] in the same way as \bert, denoted as $\bm{e}$. Then, we compute the pairwise relevance score $r_{ij}$ as follows:

\begin{align}
   & \bm{h} = \bm{W_e}\ \bm{e} + \bm{b_e} \\
   & \bm{z} = [\bm{h}; \bm{v_{ij}}] \\
   & \bm{o} = softsign(\bm{W_z} \bm{z} + \bm{b_z}) \\
   & \bm{x} = softsign(\bm{W_o} \bm{o} + \bm{b_o}) \\
   & r_{ij} = sum(\bm{x})
\end{align}
where $\bm{W_e} \in \mathbbm{R}^{d_{mid} \times d_{BERT}}$ and $\bm{b_e} \in \mathbbm{R}^{d_{mid}}$ are parameters of a linear layer to reduce the dimensionality of the BERT output; $[\bm{h};\bm{v_{ij}}]$ is the concatenation of $\bm{h}$ and the pairwise vector of handcrafted features $\bm{v_{ij}} \in \mathbbm{R}^{d_f}$; $\bm{W_z} \in \mathbbm{R}^{d_o \times d_z}$, $\bm{b_z} \in \mathbbm{R}^{d_o}$, $\bm{W_o} \in \mathbbm{R}^{d_x \times d_o}$ and $\bm{b_o} \in \mathbbm{R}^{d_x}$ are parameters of two dense layers with the  \textit{softsign} activation function; $sum(\bm{x})$ represents the sum of  values in vector $\bm{x}$. 

In \berttd. The time difference feature between $u_i$ and $u_j$ is a 6-d vector:
$$[n', x_1, x_2, x_3, x_4, x_5]$$
where $n' = (i - j) / 100$ representing the relative distance between two utterances in the candidate pool; $x_1, \cdots, x_5$ are binary values indicating whether the time difference in minutes between $u_i$ and $u_j$ lies in the ranges of $[-1,0), [0, 1), [1, 5), [5, 60)\ \mbox{and}\ (60, \infty)$ respectively.

\subsection{Pairwise Models Settings}
\paragraph{Model architecture and training}
We choose the best hyper-parameters according to the \textit{ranking} performance Recall@$1$ on validation set. All models are evaluated every 0.2 epoch. We stop training if Recall@$1$ on validation set does not improve in three evaluations consecutively. 

The final settings are as follows. In \mf, we use a 2-layer FFN with \textit{softsign} activation function. Both layers contain $512$ hidden units. We train it using Adam optimizer with learning rate $0.001$. 
For all transformer-based models (\bert, \bertmf, \albert and \polyencoder), we use Adamax optimizer with learning rate $5 \times 10^{-5}$, updating all parameters in training.  We use automatic mixed precision to reduce GPU memory consumption provided by Pytorch\footnote{https://pytorch.org/}. All experiments are implemented in Parlai\footnote{https://parl.ai/}.

\subsection{BGMCD Set Up}
\paragraph{Setup} Both node frequency estimation and graph construction are based on the relevance scores from \bertmf. In the rule-based method, we choose $\alpha$ in $\{0.9, 1,1, 1.3, 1.5, 1.7, 1.9\}$ and $\beta$ in $\{0.1, 0.2, 0.3, 0.4, 0.5\}$. The optimal values $\alpha=1.3$ and $\beta=0.2$ yield the best link prediction F1 on the validation set. The regression mode is a 2-layer fully connected neural network. Both layers contain 128 hidden units, with the ReLU activation function. We choose hidden layer size from $\{64, 128, 256\}$ and the number of layers from $\{2, 3\}$. We train the model using Adam optimizer with batch size 64. Hyper-parameters are chosen to minimize mean squared error on the validation set. The integer programming problem is solved using pywraplp\footnote{\url{https://google.github.io/or-tools/python/ortools/linear_solver/pywraplp.html}}. We observe that sometimes the integer programming problem is infeasible due to underestimation of the frequencies of some nodes. We relax Equation~\ref{eq:bgmcd} in experiments as follows to avoid infeasibility:
\begin{equation}
\begin{array}{ll@{}ll}
                & \text{max} \displaystyle\sum\limits_{\langle v_i,v_j \rangle \in E} x(i,j) \cdot w(i,j)  & \\
    \text{s.t.} &  & &\\ 
            & \displaystyle\sum\limits_{v_l \in neighbors(v_i)}   x(i,l)  \bm{\leq} 1, & \forall v_i \in V_l \\
            & \displaystyle\sum\limits_{v_p \in neighbors(v_j)}   x(p,j) \leq 1, & \forall v_j \in V_r \\
            & x(i,j) \in \{0, 1\}
\end{array}
\end{equation}

\end{document}